\documentclass[conference]{IEEEtran}
\IEEEoverridecommandlockouts
\usepackage{cite}
\usepackage{amsmath,amssymb,amsfonts}
\usepackage{algorithmic}
\usepackage{graphicx}
\usepackage{textcomp}
\usepackage{xcolor}
\usepackage{subcaption}

\usepackage{pifont}
\newcommand{\cmark}{\ding{51}}%
\newcommand{\xmark}{\ding{55}}%

\def\BibTeX{{\rm B\kern-.05em{\sc i\kern-.025em b}\kern-.08em
    T\kern-.1667em\lower.7ex\hbox{E}\kern-.125emX}}
\begin{document}

\title{
MAISON - Multimodal AI-based Sensor platform for Older Individuals



}


\author
{
\IEEEauthorblockN{
Ali Abedi\textsuperscript{1},
Faranak Dayyani\textsuperscript{1,2},
Charlene Chu\textsuperscript{1,3},
Shehroz S. Khan\textsuperscript{1,2}}

\IEEEauthorblockA{
\textsuperscript{1}KITE – Toronto Rehabilitation Institute, University Health Network, Canada\\
\textsuperscript{2}Institute of Biomedical Engineering, University of Toronto, Canada\\
\textsuperscript{3}Lawrence S. Bloomberg Faculty of Nursing, University of Toronto, Canada\\
    ali.abedi@uhn.ca,
    faranak.dayyani@mail.utoronto.ca,
    charlene.chu@utoronto.ca,
    shehroz.khan@uhn.ca}
}

\maketitle

\begin{abstract}
There is a global aging population requiring the need for the right tools that can enable older adults' greater independence and the ability to age at home, as well as assist healthcare workers. It is feasible to achieve this objective by building predictive models that assist healthcare workers in monitoring and analyzing older adults' behavioral, functional, and psychological data. To develop such models, a large amount of multimodal sensor data is typically required. In this paper, we propose MAISON, a scalable cloud-based platform of commercially available smart devices capable of collecting desired multimodal sensor data from older adults and patients living in their own homes. The MAISON platform is novel due to its ability to collect a greater variety of data modalities than the existing platforms, as well as its new features that result in seamless data collection and ease of use for older adults who may not be digitally literate. We demonstrated the feasibility of the MAISON platform with two older adults discharged home from a large rehabilitation center. The results indicate that the MAISON platform was able to collect and store sensor data in a cloud without functional glitches or performance degradation. This paper will also discuss the challenges faced during the development of the platform and data collection in the homes of the older adults. MAISON is a novel platform designed to collect multimodal data and facilitate the development of predictive models for detecting key health indicators, including social isolation, depression, and functional decline, and is feasible to use with older adults in the community.
\end{abstract}


\section{Introduction}
The global number of people aged $65$ and over is expected to double to more than $1.5$ billion in $2050$ \cite{chang2019measuring}. The aging process increases one's likelihood of developing a variety of health problems, including dementia, diabetes, cardiovascular disease, osteoarthritis, or other chronic conditions \cite{kim2022attitudes}. The World Health Organization data indicates that there was a $7.2$ million healthcare worker shortage in $2013$, and that number is expected to rise to over $12.9$ million by $2035$ \cite{turner2018no}. The discrepancy between the number of older adults who may need care and healthcare workers who can provide that care will result in essential healthcare needs that will be unmet, resulting in poorer health and well-being \cite{healthcare2022}.

Recent advancements in Internet-of-Things (IoT) provide the opportunity to collect multimodal sensor data from the homes of patients and older adults and store it in cloud servers \cite{khan2021federated}. Data from multiple sensing modalities are often required to model complex behavioral, physiological, and social behaviors of individuals living in the community, including multifaceted health issues such as social isolation, mental health issues, and frailty. Artificial Intelligence (AI) approaches can be used on this plethora of multimodal data to detect the onset of various conditions \cite{khan2017daad}, issue alerts to caregivers, and potentially provide recommendations for treatments or healthcare services. The various sensing modalities that AI models leverage include, but are not limited to, motion from an accelerometer, heart rate from a photoplethysmography (PPG) sensor, images/videos captured from cameras, ambient light levels, motion sensors, and smartphone (phone) ambient sound recordings \cite{au2022monitoring}.

It is challenging to collect multimodal data from multiple devices located in different locations to train AI models. Oftentimes, researchers have to manually combine information from different modalities into a single database \cite{khan2017daad}. In order to meet their specific needs, researchers sometimes develop their own smart sensors; however, this bespoke development potentially increases the costs of projects. In order to mitigate these limitations, researchers have looked towards commercially available smart devices. One of the challenges in using these off-the-shelf devices is limited interoperability because each has its own Application Programming Interface (API) and private cloud that limits researcher access to the collected data. For example, Apple products \cite{hernando2018validation} use their own protected communication protocols, making access to the raw data impossible. Another issue with commercial sensors and devices is that, in most cases, the devices provide processed sensor data (e.g., average heart rate and the number of steps), which may not be ideal for developing predictive models. From a machine learning standpoint, raw sensor data can be used to extract informative features required for training machine learning \cite{spasojevic2021pilot,khan2017detecting,kim2021machine,chikersal2021detecting} or learn features using deep learning methods \cite{khan2017detecting,nogas2020deepfall,nguyen2021federated}. More recently, many sensor manufacturers provide access to raw sensor data through their APIs \cite{withings2022,insteon2022,smartthings2022} that enables the development of new cloud-based IoT platforms.

There are currently a number of existing cloud-based digital health platforms that integrate various sensing devices to provide backend infrastructure and management interfaces \cite{far2021jtrack,ferreira2015aware,ranjan2019radar,asareme,torous2016new,wang2021hopes}. However, very few of them provide comprehensive functionality and often lack essential features that play a major role in preventing gaps in the data and battery depletion, including location geofencing and compatibility with Wear Operating System (OS) for smartwatches (watches). In order to take advantage of the potential of different types of commercial sensors, our goal is to develop a scalable cloud-based digital health platform that is able to provide a holistic perspective of active aging in the home by (i) seamlessly collecting data from multiple smart devices from different manufacturers and (ii) storing the data in an organized manner in the cloud to facilitate the development of predictive models that support well-being. In this paper, we introduce MAISON (Multimodal AI-based Sensor platform for Older iNdividuals), a novel digital health platform for securely collecting raw sensor data from various devices, including smartphones, smartwatches, and non-wearable ambient sensors. We deployed MAISON in the homes of two older adults living in the community with post-hip surgery for a period of two months each and successfully collected multimodal sensor data. The following sections describe the related work, details of the MAISON platform, data collection, and preliminary data analysis.

\section{Related Work}
\label{related_work}
In this section, we present a brief overview of previous digital health platforms that include phones, watches, and IoT sensors. We limit our description to focus on the research and development work of each platform rather than their predictive or analytics aspects.

Khan et al. \cite{khan2017daad} presented DAAD, a multimodal sensing framework to detect incidences of agitation and aggression in people with dementia, using multiple video cameras, an Empatica E4 wristband, a pressure mat, and motion sensors to collect various types of data from people living with dementia. The Empatica E4 wristband continuously collected data and stored the information in its internal memory, from where it was periodically pushed to its own cloud. The DAAD system does not use a central server to direct various data from various modalities automatically. A research assistant was required to manually download the data from the cloud on a daily basis for further analysis. For a long-term research study or lengthy data collection period, this can be very time-consuming and lead to cost and resource overruns. Ranjan et al. \cite{ranjan2019radar} developed RADAR-base (Remote Assessment of Disease and Relapse) that can collect, monitor, and analyze data via wearable and mobile devices. RADAR-base does not utilize external IoT devices. Ferreira et al. \cite{ferreira2015aware} developed AWARE, a general-purpose Android phone application designed to "capture, infer, and provide human-based contextual information on mobile devices". The major limitation of AWARE is that it does not support other endpoint devices (e.g., watches and motion detectors) outside of phones. Kennedy et al. \cite{asareme} developed Me, a mobile application that collects phone sensor data (location, physical activity, and light) and self-reported surveys to detect the onset of depression. Since Me is built on AWARE, it does not support wearable devices or external sensors. Torous et al. \cite{torous2016new} reported on the development and initial utilization of Beiwe that can collect location data, accelerometer, phone calls and text messages, audio recording, and screen and phone status. Wang et al. \cite{wang2021hopes} developed HOPES (Health Outcomes through Positive Engagement and Self-Empowerment) for the collection of data from phones and wearable devices (Fitbit watch). Based on the Beiwe platform, HOPES expands data collection to include wearable devices and additional phone sensors. Neither the Beiwe nor the HOPES support third-party external ambient sensors. Far et al. \cite{far2021jtrack} developed JTrack, an Android phone application to collect in-built phone data, and a server-side dashboard to manage and create studies. JTrack is not equipped to connect and collect data from any other device besides Android phones.

Table I provides a comparison of MAISON with other digital health platforms. Besides DAAD \cite{khan2017daad}, which does not offer a cloud server for automatic data collection and relies on manual data transfer from sensors to computers, none of the available platforms are able to collect data from external ambient sensors, such as motion sensors and sleep tracking devices \cite{withings2022,insteon2022,smartthings2022}. In addition, none of the available platforms is developed to integrate and connect the Google Wear OS watches with Android phones. The MAISON system addresses all the above features to continuously collect data from participants in their homes with zero effort.

\begin{table}[t]
\centering
\caption{
Comparing DAAD \cite{khan2017daad},
RADAR-base \cite{ranjan2019radar},
AWARE \cite{ferreira2015aware},
Me \cite{asareme},
Beiwe \cite{torous2016new},
HOPES \cite{wang2021hopes},
JTrack \cite{far2021jtrack},
and our proposed platform, MAISON (described in Section \ref{MAISON}).}
\resizebox{\columnwidth}{!}{
\begin{tabular}{l c c c c c}
 \hline
 Platform & Phone & Watch & third-party & Cloud & Geofencing \\ [0.5ex]
  & support & support & sensors & backend & feature\\ [0.5ex]
 \hline\hline
 DAAD \cite{khan2017daad} & \cmark & \cmark & \cmark & \xmark & \xmark \\ 
 RADAR-base \cite{ranjan2019radar} & \cmark & \cmark & \xmark & \cmark & \xmark \\ 
 AWARE \cite{ferreira2015aware} & \cmark & \xmark & \xmark & \cmark & \xmark \\ 
 Me \cite{asareme} & \cmark & \xmark & \xmark & \cmark & \xmark \\ 
 Beiwe \cite{torous2016new} & \cmark & \xmark & \xmark & \cmark & \xmark \\ 
 HOPES \cite{wang2021hopes} & \cmark & \cmark & \xmark & \cmark & \xmark \\ 
 JTrack \cite{far2021jtrack} & \cmark & \xmark & \xmark & \cmark & \xmark \\ 
 \textbf{MAISON (proposed)} & \cmark & \cmark & \cmark & \cmark & \cmark \\
 \hline
\end{tabular}
}
\label{table:related_work}
\end{table}

\section{MAISON Platform}
\label{MAISON}
An overview of the MAISON digital health platform is shown in Figure \ref{fig:block_diagram}, which consists of four main components: (1) the MAISON phone application, (2) the MAISON watch application, (3) non-wearable sensors and their associated clouds, (4) the MAISON central cloud.

\subsection{Smartphone Application}
\label{smartphone}
\begin{figure}[t]
    \centering
    \includegraphics[scale=0.1367]{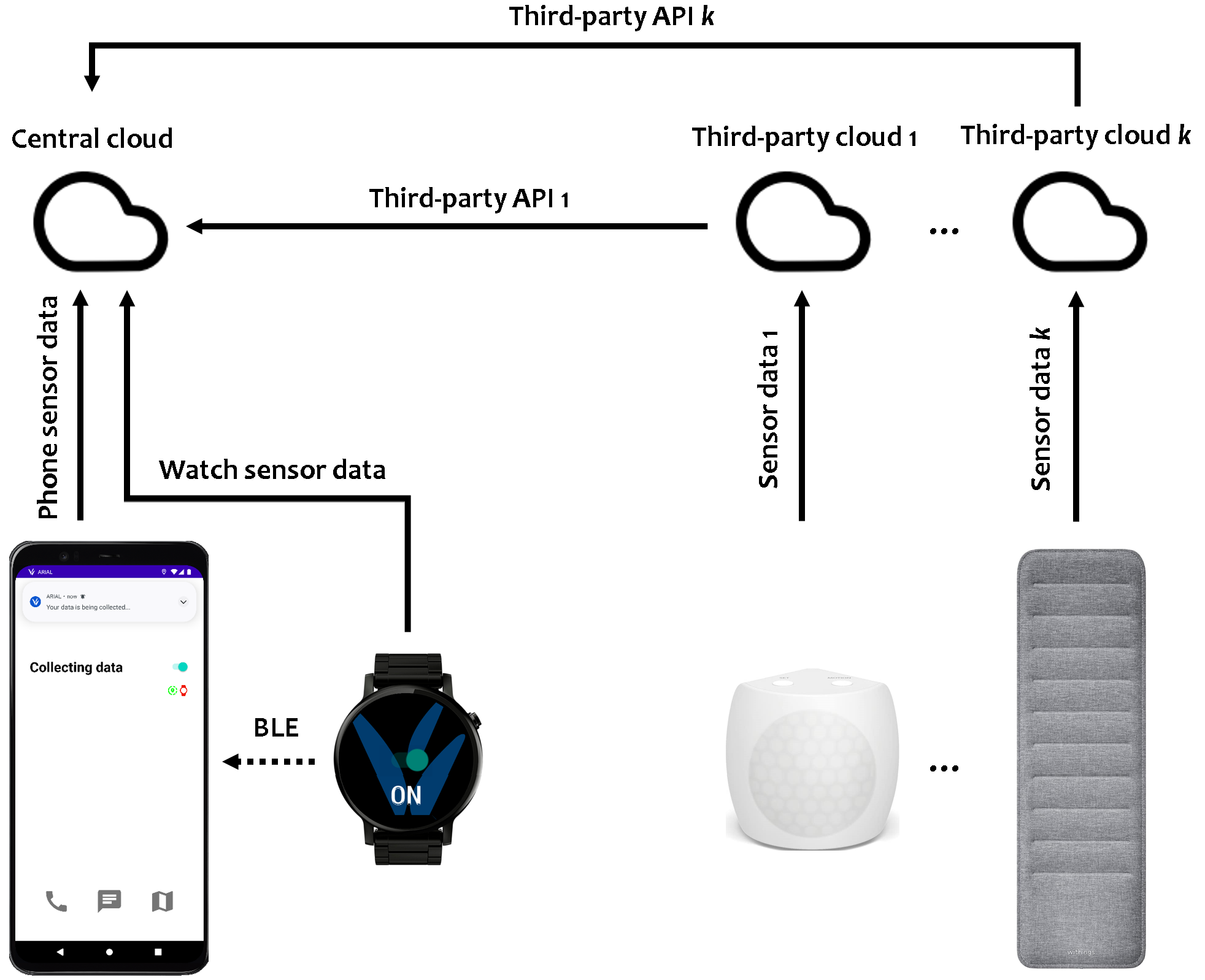} \\
    \caption{Block diagram of the MAISON platform for collecting data from phones, watches, and external sensors, see Section \ref{MAISON}.
    }
    \label{fig:block_diagram}
    
\end{figure}

The smartphone application is developed to (1) collect and store sensor data using phone built-in sensors, (2) establish a connection with the watch to receive and temporarily store data collected by the watch sensors, and (3) send the collected data to the central cloud. The main Android activities of the phone and watch applications are shown in Figure \ref{fig:block_diagram}. The application has a simple and straightforward design to improve usability, which is an important consideration for older adult end-users. Consented participants are empowered with the ability to stop and start MAISON’s data collection when they choose. Both the phone and watch applications are programmed to be launched and start data collection automatically after each reboot of the devices to increase ease of use for older adults.

\subsubsection{Smartphone Data Collection}
\label{smartphone_sensor_data_collection}
Android phones have various built-in sensors -- GPS, accelerometer, light, and other sensors that provide information on position, motion, and environment. Developers have access to both the raw and processed data generated by the sensors. MAISON can include any of the Android phone's built-in sensors as part of the data collection and storage on the cloud. Sampling frequency can be determined for each sensor depending on the study requirements. MAISON uses Google's Fused Location Provider API to collect location data. In order to avoid rapid battery drain due to continuous location data collection, MAISON uses Geofencing, which leverages GPS signals to identify a participant's location \cite{geofencing2022}. A perimeter can be created using the Geofencing feature that can surround the house of the participant, and data collection only triggers when the participants leave the geofence. All the sensor data collected on the phone is inserted into its local database in real-time and periodically sent to the cloud.

\subsubsection{Smartphone and Watch Connectivity}
Embedded sensors in the smartwatch can collect physiological and activity data in the MAISON platform (see Section \ref{watch}). The watch sends the collected data directly to the cloud. When the watch is not connected to the Internet, it sends the collected data to the phone using Bluetooth Low Energy (BLE) at specific intervals. Contrary to the previous works (see Section \ref{related_work}), in MAISON, both the watch and phone operate on Google Android and Wear OS that are fully compatible, and enable constant connectivity between them using BLE. Once the phone receives data from the watch, it temporarily stores the data in its local database and then sends the data to the cloud.

\subsubsection{Data Transfer to the Cloud}
\label{smartphone_data_transfer}
Data stored in the local database of the phone application is sent to the MAISON central cloud (Google Firebase Cloud Firestore) using secure Hypertext Transfer Protocol Secure (see Section \ref{cloud}) at specified time intervals as per the study design. Once the phone receives an acknowledgement that data has been successfully stored in the cloud, it permanently deletes the data from its local database. At these timestamps, the phone also checks its local database to determine whether any data has been received from the watch in the past time interval. In case of no data from the watch, the phone displays a notification to instruct the participant to wear the watch and turn on the watch application.

\subsection{Smartwatch application}
\label{watch}
The phone and the watch collect complementary information. The software architecture of the MAISON Wear OS watch application is similar to that of the Android phone application, with a few subtle differences. At predetermined time intervals, the watch attempts to transmit data to the cloud. In case the watch is not connected to the Internet, it will send its collected data via BLE to the phone, which will be transferred to the cloud afterward.

\subsection{Ambient Sensors and Third-party Clouds}
\label{external}
The MAISON platform currently supports the Insteon motion sensors \cite{insteon2022} and Withings sleep sensors \cite{withings2022}. Using the manufacturer provided APIs and secure authentication, the data from these devices is collected from their clouds and stored on the MAISON central cloud. These sensors can be added or removed as per a study's requirement.

\subsection{MAISON central cloud}
\label{cloud}
As per research ethics requirements, MAISON's multimodal sensor data is to be stored on Google Cloud servers located in Canada. All the recommended practices by Google Cloud are taken into consideration in MAISON central cloud development to meet the HIPAA requirements \cite{cloud2022}. The collected data modalities are anonymously transferred to the MAISON central cloud and saved as CSV files. In the central cloud, a data merging algorithm is deployed to merge data received from the phone and watch at different times and output one CSV file for each data modality at each time segment, e.g., an hour or a day. In addition, we developed Google Cloud Functions to securely authenticate through OAuth 2.0 and access the participants' multimodal sensor data on the third-party clouds (see section \ref{external}), and transfer data from those clouds to the MAISON central cloud.

\begin{figure}[]
    \centering
        \begin{subfigure}{0.35\textwidth}
            \includegraphics[scale=.22]{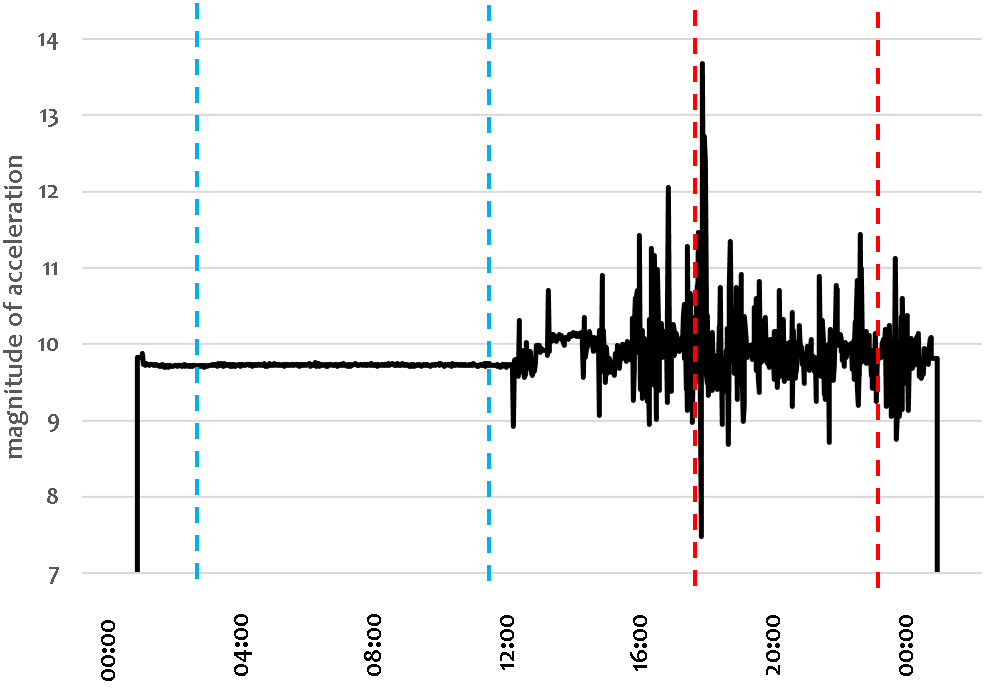}
            \caption{}
            \label{fig:step}
        \end{subfigure}
        \begin{subfigure}{0.35\textwidth}
            \includegraphics[scale=.22]{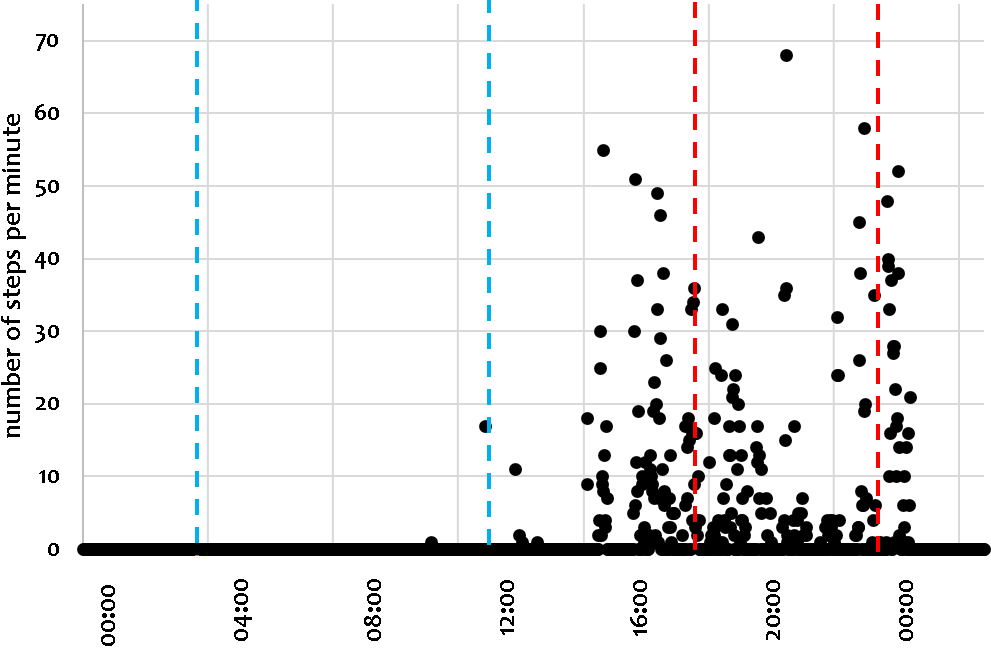}
            \caption{}
            \label{fig:step}
        \end{subfigure}
        \begin{subfigure}{0.35\textwidth}
            \includegraphics[scale=.22]{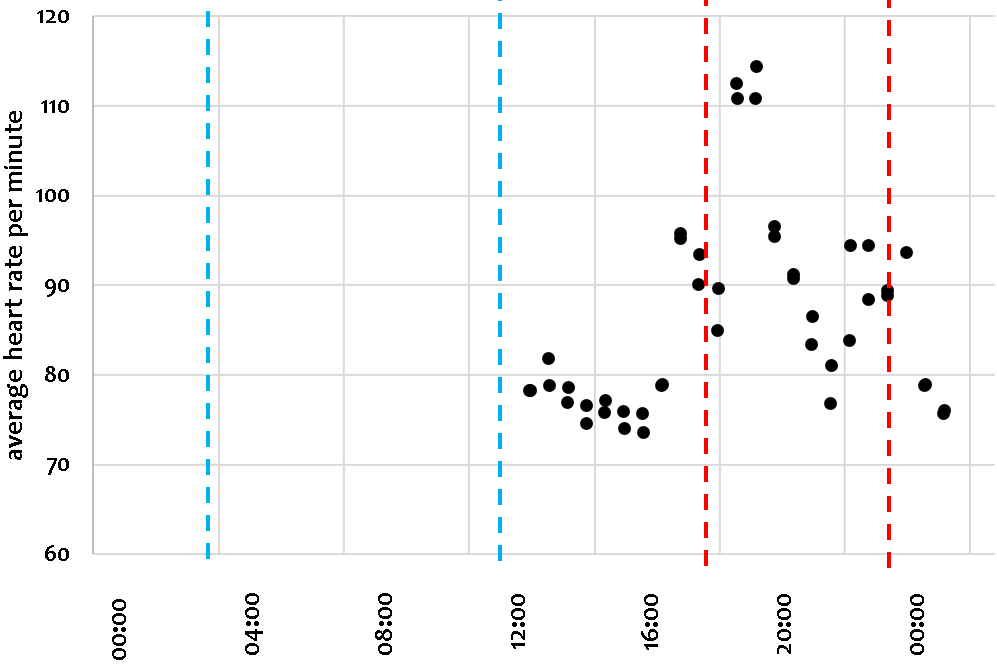}
            \caption{}
            \label{fig:step}
        \end{subfigure}
        \begin{subfigure}{0.35\textwidth}
            \includegraphics[scale=.22]{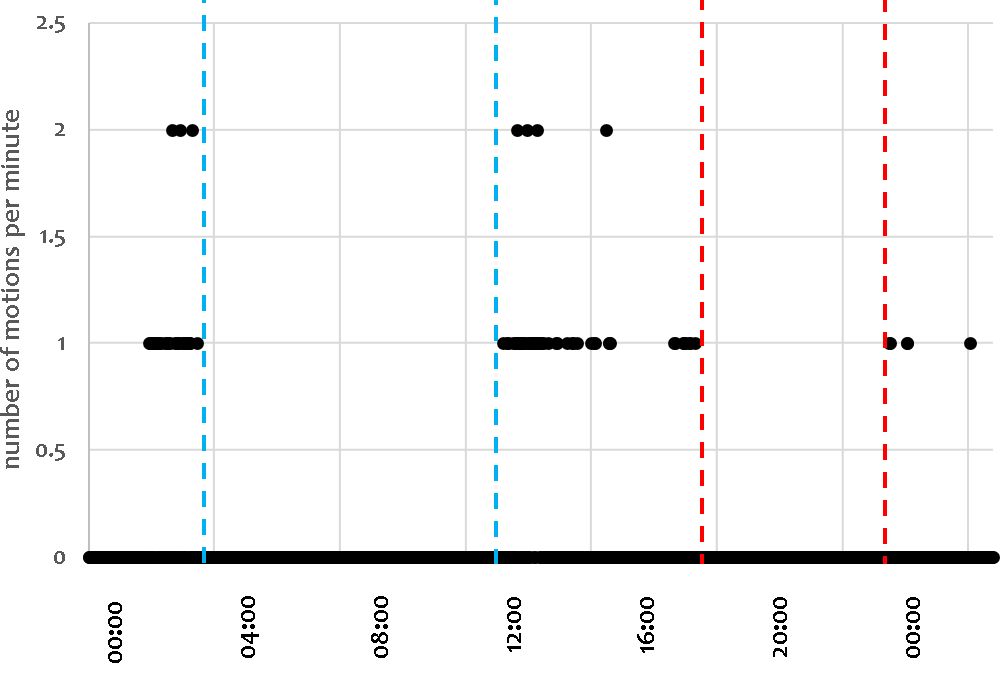}
            \caption{}
            \label{fig:step}
        \end{subfigure}
        \begin{subfigure}{0.3\textwidth}
            \includegraphics[scale=.2]{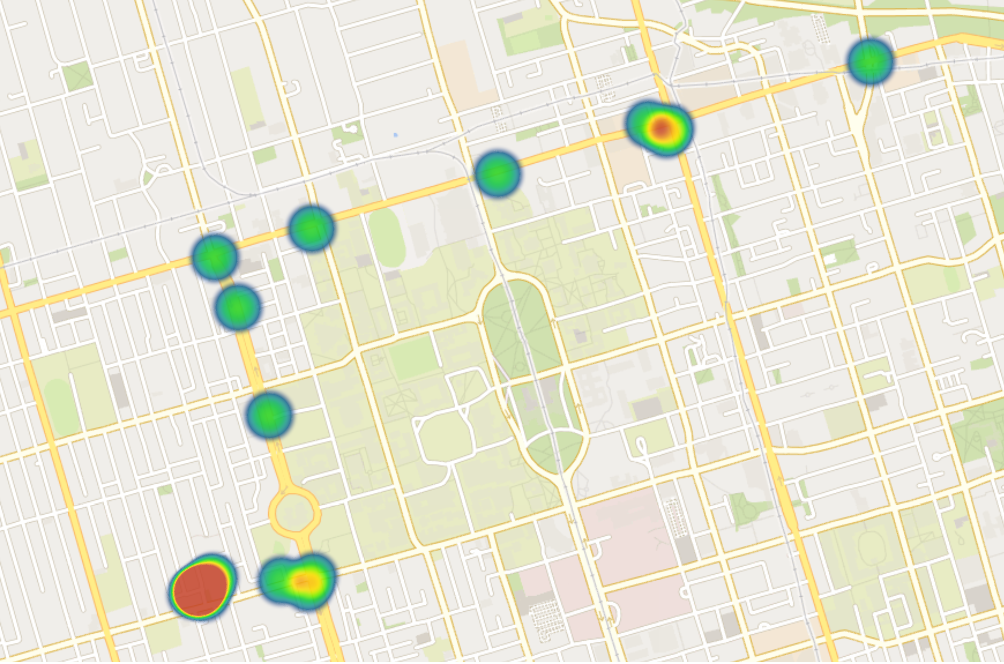}
            \caption{}
            \label{fig:step}
        \end{subfigure}        
    \caption{
    (a-e) A visual representation of different sensor data modalities collected from one of the older adult participants (described in Section \ref{experiments}) over a period of $24$ hours.
    }
    \label{fig:one_participant}
\end{figure}

\section{Experiments}
\label{experiments}
The MAISON platform was deployed into the homes of two older adult patients who were discharged home post-hip fracture surgery, and we collected data for eight weeks. Both the patients were females in their 70s, who had recently undergone hip fracture surgery, lived alone with no pets, and had WiFi in their homes. The MAISON phone, and watch applications were installed on two Motorola Moto G7 Android phones, and two Mobvoi TicWatch Pro 2020 Wear OS watches, respectively. The Withings sleep tracking mattress, and the Insteon motion sensor were installed underneath the bed mattress, and in the living rooms of the participants' homes, respectively. Bi-weekly clinical questionnaires were also collected, including the Social Isolation Scale \cite{ranjan2019social} and Oxford hip score \cite{wylde2005oxford}. This is an ongoing study, and the overall goal is to leverage the MAISON platform to build predictive models from this multimodal sensor data.
The following multimodal sensor data were collected from the two participants:

\begin{itemize}
    \item On the phone, location GPS data at a frequency of 0.033 Hz with the house of each participant as the geofence.
    
    \item On the watch, the accelerometer data at a frequency of $1$ Hz, the PPG heart rate sensor was activated every $30$ minutes and collected data for $30$ seconds, and number of steps collected by the step detector of the watch.
    
    \item  The timestamp of motion events is collected from the Insteon Motion sensor \cite{insteon2022}. 
    
    \item Sleep data using Withings mattress at each nap and night sleep, including total sleep duration, deep sleep duration, average heart rate, and snoring time \cite{withings2022}.

\end{itemize}

Figure \ref{fig:one_participant} shows
(a) the magnitude of the watch accelerometer,
(b) the number of steps per minute detected by the watch,
(c) the average value of the watch heart rate every minute,
(d) the total number of motion sensor events every minute, along with the sleep time (the time period between the two blue dashed lines) of one of the participants for $24$ hours. It also shows (e) the latitude and longitude coordinates of the location data of the participant on a map while the participant is outside geofence (the time period between the two red dashed lines). Heat map colours in Figure \ref{fig:one_participant} (e) represent the frequency of occurrence of location data at each location.

The complementary information of the collected data can be observed in Figure \ref{fig:one_participant}. The sleep began at approximately 02:30 and ended at 11:30, between the two blue dashed lines. There were no motion events during the sleep period. After 11:30, motion events began occurring until around 18:30, when the participant left geofence. The location data collection began at 18:30 and ended at 22:30 when the participant re-entered geofence. Motion data collection started again at this time, 22:30. During the wake time of the participant, acceleration, heart rate, and step data were continuously collected, except when the watch was being charged or when it was not worn.

\section{Discussion}
\label{discussion}
During the development of the system, we encountered minor practical challenges. One technical challenge was that the regular updates to the Android OS and its dependencies led to modifying the code of the phone and watch applications to enable running on different Android devices.
Another challenge was frequent updates in the APIs of external non-wearable ambient sensors leading to code modifications.
A variety of strategies were used to optimize the performance of the phone and watch applications in order to minimize their battery consumption. The Google Geofencing API is used in conjunction with the location API to collect the location data only when the participants are outside their homes (geofences). Even though the phone application does not greatly affect the battery consumption of the phone, we found that the MAISON watch application resulted in shorter battery life of the watch.
Since participants needed to recharge the watch for approximately one and a half hours, there were gaps in the data collected from the watch during this period. Some participants did not wear the watch on certain occasions, despite the periodic notification sent on the watch. One participant highlighted the discomfort of wearing the watch.
Additionally, motion sensor and sleep mattress data could be noisy if other people are present alongside the study participant. Lastly, a participant may unplug the external sensors mistakenly, resulting in missing data.


\section{Conclusion and Future Work}
\label{conclusion}
MAISON is a feasible digital health platform that can collect multimodal sensor data from various wearable and ambient devices to provide a holistic perspective of active aging. To date, it has been employed in the homes of two older adults with minor issues. The system usability \cite{lewis2018system} by older adults living in the community is being evaluated in an ongoing research study. As MAISON is easy to use, it will appeal to both older adults and their caregivers who may be younger. Significantly, MAISON could lead to an entire cloud pipeline from data collection to machine-learning and deep-learning model building to further speed up research. In future iterations, MAISON will be customized to add different clinical questionnaires or other types of wearable watches, phones, and external ambient sensors \cite{withings2022,smartthings2022,insteon2022}.

\end{document}